\documentclass[preprint,authoryear,12pt]{elsarticle}

\usepackage{lineno}
\usepackage{graphicx} 
\usepackage{epstopdf}
\usepackage{stfloats}
\usepackage{bm}
\usepackage{amssymb}
\usepackage{subcaption}
\usepackage{multirow}
\usepackage{amsmath}
\usepackage{tikz}
\usepackage{booktabs} 
\usepackage{algorithm} 
\usepackage{algpseudocode} 
\usetikzlibrary{mindmap,backgrounds}
\usetikzlibrary{shapes,arrows,patterns} 

\begin{document}
\begin{frontmatter}

\title{Graph Neural Network-Based Semi-Supervised Open-Set Fault Diagnosis for Marine Machinery Systems}
\cortext[cor1]{Corresponding author}
\author[label1]{Chuyue Lou \corref{cor1}}
\ead{lcy2024@zjvtit.edu.cn}
\author[label3]{M. Amine Atoui}
\author[label1]{Xiaomin Zhang}
\author[label1]{Dong Xu}
\author[label1]{Haijun Zhong}
\address[label1]{Intelligent Transportation School, Zhejiang Institute of Communications, China}
\address[label3]{Center for Applied Intelligent Systems Research, Halmstad University, Sweden}

\begin{abstract}
	Recently, fault diagnosis methods for marine machinery systems based on deep learning models have attracted considerable attention in the shipping industry. Most existing studies assume fault classes are consistent and known between the training and test datasets, and these methods perform well under controlled environment.  In practice, however, previously unseen or unknown fault types (i.e., out-of-distribution or open-set observations not present during training) can occur, causing such methods to fail and posing a significant challenge to their widespread industrial deployment. To address this challenge, this paper proposes a semi-supervised open-set fault diagnosis (SOFD) framework that enhances and extends the applicability of deep learning models in open-set fault diagnosis scenarios. The framework includes a reliability subset construction process, which uses a multi-layer fusion feature representation extracted by a supervised feature learning model to select an unlabeled test subset. The labeled training set and pseudo-labeled test subset are then fed into a semi-supervised diagnosis model to learn discriminative features for each class, enabling accurate classification of known faults and effective detection of unknown samples. Experimental results on a public maritime benchmark dataset demonstrate the effectiveness and superiority of the proposed SOFD framework.
\end{abstract}

\begin{keyword}
	Open-set, deep learning, fault diagnosis, marine machinery system
\end{keyword}
\end{frontmatter}

\section{Introduction}
\label{sec:introduction}

Ships are vital to international trade and maritime transportation, and ensuring their safety and reliability is one of the most critical tasks. In the past decade, more than 26,000 shipping incidents have been reported worldwide, resulting in serious economic losses and fatalities \citep{xu2023fault}. Among them, about one-third of the incidents were caused by marine machinery faults \citep{lazakis2018predicting}. Traditional marine machinery monitoring is largely dominated by human cognition and experience, such as reactive maintenance and regular inspections \citep{raptodimos2020application}. However, with the improvement of ship automation, marine mechanical systems are becoming increasingly complex, which brings challenges to fault diagnosis. 

Currently, the data collected from marine machinery systems is showing a rapid growth trend, making data-driven intelligent fault diagnosis methods an emerging research topic in the shipping industry. As an key component of data-driven methods, signal processing extracts time domain and frequency domain information related to system status from marine machinery signals, providing an effective technology for fault diagnosis \citep{zhao2023variational,li2023refined}. Machine learning another core branch of data-driven methods, includes Support Vector Machine, which are among the most commonly employed methods for fault diagnosis \citep{tan2021multi, vanem2021unsupervised}. More recently, deep learning has enabled adaptive feature extraction and end-to-end representation of complex systems, achieving notable progress in practical applications \citep{lv2024fault, li2025dual}.mong these approaches, models such as Convolutional Neural Networks (CNNs), Long Short-Term Memory (LSTM) networks, and Variational Autoencoders (VAEs) have received significant attention. Han et al. introduced a CNN with control signals and logged ship motions as input to detect and isolate thruster failures for dynamically positioned vessels \citep{han2020deep}. Shahid et al. preprocessed sensor signals to obtain crank angle signals and then trained a CNN to detect cylinder misfires and engine load conditions in a multi-cylinder internal combustion diesel engine. \citep{shahid2022real}. Xu et al. integrated the multi-head attention mechanism, convolutional layers, and residual structures, which have the ability to extract multi-scale fault features and can also capture the global and local fault characteristics of onboard sensor signals \citep{xu2023fault}. Wang et al. innovatively proposed a random CNN network architecture for diesel engine health monitoring that has all the advantages of deep learning and ensemble learning \citep{wang2021random}. Gao et al. proposed a fault diagnosis method based on improved LSTM neural network with beluga optimization algorithm, which provided a more precise solution for marine diesel engine piston ring fault diagnosis \citep{gao2024marine}. Han et al. used LSTM to build a fault prediction model to predict the remaining useful life of a ship diesel engine based on sensor measurements under two different engine load profiles \citep{han2021faulta}. Ellefsen et al. trained a VAE to estimate velocity and acceleration calculations of the anomaly score, while establishing both generic and dynamic threshold limits to detect faulty time steps online \citep{ellefsen2020online}. Velasco-Gallego et al. combined LSTM and VAE for anomaly detection performance in tandem with image generation throughstatistical analysis methods \citep{velasco2022radis}. The above literature indicates that deep learning-based fault diagnosis is a rapidly developing field. End-to-end network framework formed by the deep learning can simultaneously perform the feature extraction and fault diagnosis tasks. Nonlinear input data can be transformed into abstract feature representations and directly used to optimize the final task objectives.

It is worth noting that data-driven fault diagnosis methods perform well under ideal scenarios where sufficient labeled data and classes are available. However, faulty operating states of marine mechanical systems last for a short time, leading to limited labeled data, which in turn increases the difficulty of diagnosis. Several scholars have proposed deep learning solutions to improve the fault diagnosis capabilities of marine machinery with insufficient labeled data. Han et al. proposed a semi-supervised LSTM-VAE, which only requires fault-free data collected from the diesel engine of the research vessel Gunnerus for training \citep{han2021faultb}. Wang et al. trained a 1-D CNN with a new representation of the input signal composed by data augmentation by truncation and adjacent sample augmentation, and then used limited labeled data for fine-tuning \citep{wang2023self}. Wang et al. explored the similarities of linear and nonlinear relationships between samples, converted the training data into distance topological graphs and probability topological graphs, and performed graph learning to provide deeper information for fault diagnosis \citep{wang2022dpgcn}. The above literature is based on the closed set assumption that fault classes are known a priori, which restricts diagnosis models to make decisions within a limited number of known classes. 

In the field of industrial machinery, open-set fault diagnosis has emerged as an urgent challenge. Discriminative methods have become the mainstream solution, as they enhance the discriminability of network representations during training so as to correctly detect samples from unknown class samples \citep{zhou2024contrastive}. Guo et al. introduced a composite transferability metric to differentiate unknown fault samples from labeled known classes, and deployed a parameter-adaptive unsupervised transfer algorithm to discern the count of new fault types \citep{guo2024universal}. Wang et al. introduced self-supervised contrastive learning to extract robust discriminative features and designed a squeeze confidence rule to improve distinguishability \citep{wang2024self}. Zhao et al. used triplet loss to improve intra-class compactness and inter-class separability, and proposed a class-wise decision boundary mechanism to isolate unknown faults \citep{zhao2022adaptive}. Sun et al. developed an open set diagnosis method for rolling bearings based on a prototype and reconstructed integrated network, and designed an open set performance evaluation index to exclude unknown classes \citep{9953149}. Liu et al. extracted multi-scale time-frequency features of faults based on the multibranch convolution structure and effectively distinguished unknown faults using differences in core regions \citep{10198472}. Fu et al. constructed a multi-hop attention graph variational autoencoder to adaptively extract hierarchical features, and designed structural representation constraints and an open set diagnosis strategy based on relative distance \citep{10214203}. Notably, the performance of these open-set fault diagnosis methods is strongly influenced by the feature extraction capabilities of deep learning models. Since the learned features are optimized only for the known classes seen during training, information relevant to distinguishing unknown classes is often overlooked. As a result, models may assign high confidence to unknown samples that share similar discriminative features with known classes.

For open set recognition of visual tasks, the transductive learning strategy has been demonstrated to be effective in learning models using training and test samples \citep{yang2021s2osc,sun2024conditional}. Generally, transductive learning first provides pseudo labels for test samples with high confidence scores using a baseline model. Then, the baseline model can be updated and re-trained by jointly utilizing labeled training set and the subset of pseudo-labeled test set. However, some test samples are easily assigned to wrong pseudo labels during the transduction process, which leads to poor classification results after participating in training. Therefore, how to select a relatively reliable subset with pseudo-labels from the entire unlabeled test samples is an important issue for optimizing the performance of open-set classification. 

Inspired by transductive learning, we propose a semi-supervised open-set fault diagnosis framework, termed SOFD, which is capable of jointly detecting both known and unknown fault classes. The SOFD framework consists of three main components: supervised feature learning, reliability subset construction, and semi-supervised diagnosis. First, a Graph Convolutional Network (GCN) is employed as the feature learning model, as it performs well in capturing the complex topological relationships of data collected from marine machinery systems. The GCN is trained on all labeled training samples to obtain discriminative features that distinguish known classes. In principle, other deep learning–based fault diagnosis models could also be used for feature learning. Next, unlabeled test samples are classified using a discriminant score function in the multi-layer feature fusion space of the trained GCN. Test samples that are excluded from known classes but remain consistent with multiple neighbors in the feature space form a relatively reliable subset, which is assigned pseudo-labels of unknown classes. Finally, both the original training set and the reliable test subset are used to train a semi-supervised model, which produces label predictions for the test samples. The main contributions of this paper are summarized as follows:

\begin{enumerate}[(1)]
\item An open-set fault diagnosis framework with semi-supervised learning is designed. Pseudo-labeled test samples are involved in the training stage, which improves the discriminative ability of feature representations learned by CGN for known and unknown samples.
\item To select a reliable test subset, a new feature space of multi-layer feature fusion is constructed which can provide more fault detail information. Based on the calculated discriminant scores from statistical analysis as exclusion evidence, test samples whose pseudo labels are consistent with their neighbors are selected.
\item The effectiveness and superiority of the proposed TOSFD are demonstrated on a public dataset.
\end{enumerate}

The rest of the paper is organized as follows. Section \ref{sec:preliminaries} provides the problem statement and the basic model. Section \ref{sec:methodology} elaborates on the proposed framework. Section \ref{sec:casestudy} reports the experimental results of the proposed method. Section \ref{sec:conclusion} concludes the paper.

\section{Preliminaries}
\label{sec:preliminaries}

This section begins by presenting the case study, a formal definition of the open set-based fault diagnosis task for a marine mechanical system. We then briefly review the GCN model for feature extraction and classification.

\subsection{Propulsion system description}
\begin{figure}[h]
\centering
\includegraphics[width=0.6\hsize]{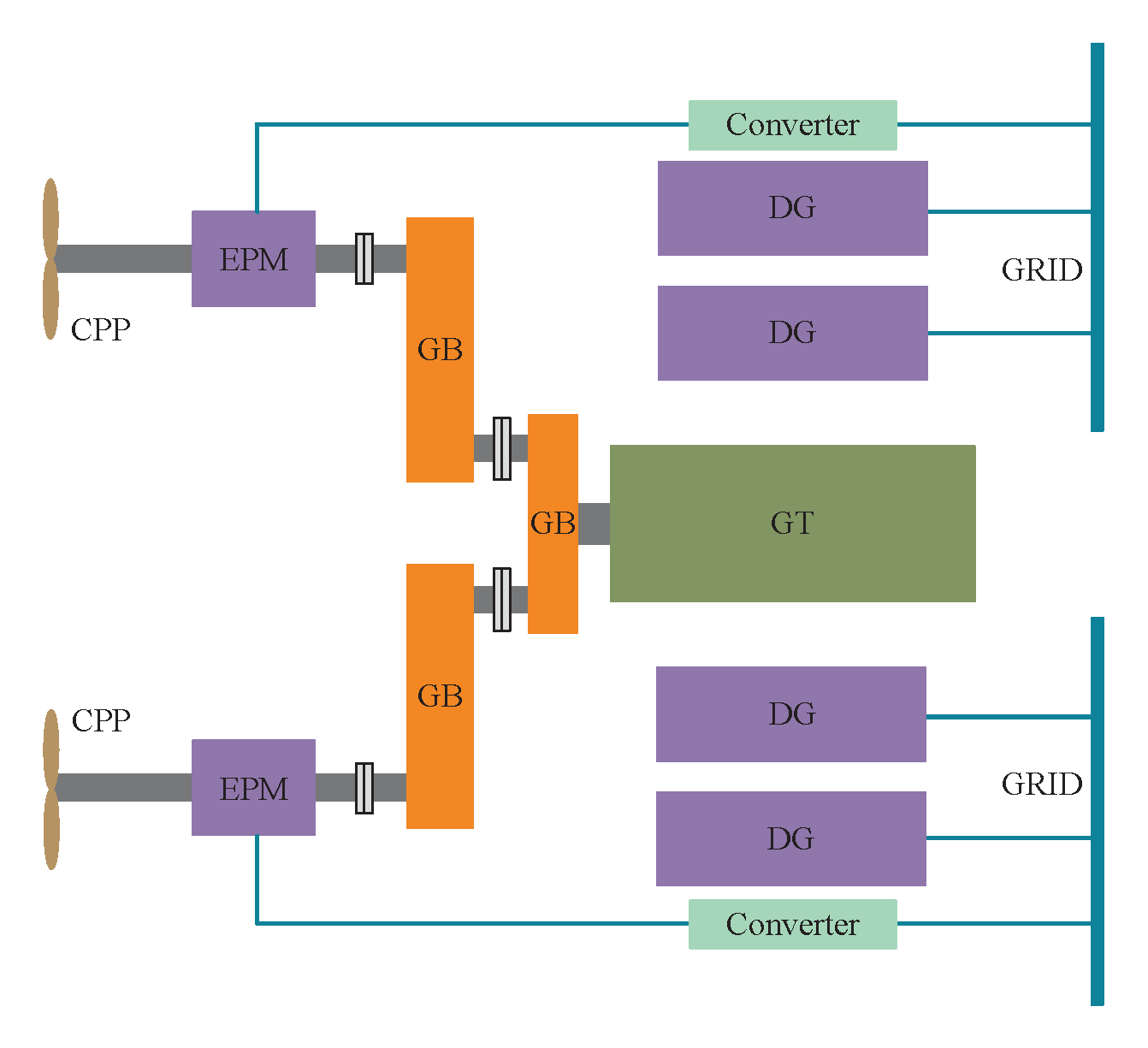}
\caption{Schematic diagram of the propulsion system \citep{tan2021multi}}.
\label{psfigure}
\end{figure}

The experiments were performed on a public dataset acquired by a complex numerical simulator of a naval propulsion system, which has been validated with real data \citep{cipollini2018condition}. The propulsion system includes a gas turbine propulsion system and an electric propulsion system, and its schematic diagram is shown in Figure \ref{psfigure}. In the propulsion system, the gas turbine mechanically drives the propeller through the transmission action of the gearbox, clutch and shaft system. The propulsion motor is installed on the transmission shaft and can also directly drive the propeller. 

The decay data of the important components of the propulsion system including a vessel Gas Turbine (GT), Gas Turbine Compressor (GTC), Hull (HLL) and Propeller (PRP) were collected at nine different speeds. The details of the normal operating condition and the four decay faults are described in Table \ref{fault}. Among the 25 sensor measurements in the dataset, the variables after removing the linearly related attributes are shown in Table \ref{dataps} \citep{carrega2019simple}.

\begin{table}[h]
\centering
\caption{Description of Continuous Process Variables.}
\label{fault}
\begin{tabular}{p{50pt}p{80pt}p{180pt}p{60pt}}
	\hline
	Condition &   Description &  Coefficients  & Number \\ \hline
	Normal    &   Normal     & $kK_t \in [0.95,1],kH \in [1,1.1]$ & 1800 \\ 
	&              & $kK_c \in [0.98,1],kM_t \in [0.99,1]$ &      \\
	Fault 1   &   PRP decay  & $kK_t \in [0.9,0.95),kH \in [1,1.1]$ & 1800 \\ 
	&              & $kK_c \in [0.98,1],kM_t \in [0.99,1]$ &      \\
	Fault 2   &   HLL decay  & $kK_t \in [0.95,1],kH \in (1.1,1.2]$ & 1800 \\ 
	&              & $kK_c \in [0.98,1],kM_t \in [0.99,1]$ &    \\
	Fault 3   &   GTC decay  & $kK_t \in [0.95,1],kH \in [1,1.1]$ & 1800 \\
	&              & $kK_c \in [0.95,0.98),kM_t \in [0.99,1]$ &      \\
	Fault 4   &   GT decay   & $kK_t \in [0.95,1],kH \in [1,1.1]$ & 1800 \\
	&              & $kK_c \in [0.98,1],kM_t \in [0.975,0.99]$ &      \\
	\hline
\end{tabular}
\end{table}

\begin{table}[h]
\centering
\caption{Description of Continuous Process Variables.}
\label{dataps}
\begin{tabular}{p{25pt}p{140pt}p{25pt}p{140pt}}
	\hline
	Order & Name & Order & Name \\ \hline
	1 & GT shaft torque  & 10 & External pressure \\ 
	2 & GT speed  & 11 & HP turbine exit pressure \\ 
	3 & Shaft torque stbd  & 12 & TCS TIC control signal \\ 
	4 & HP turbine exit temperature  & 13 & Average controllable pitch propeller thrust \\
	5 & Generator of gas speed & 14 & Average shaft rpm \\
	6 & Fuel flow & 15 & Average thrust coefficient \\
	7 & ABB TIC control signal & 16 & Average propeller rps \\
	8 & GT compressor outlet air pressure & 17 & Average propeller torque \\
	9 & CGT compressor outlet air temperature &    &   \\
	\hline
\end{tabular}
\end{table}

Within the domain of this propulsion system, several research initiatives have been undertaken for fault diagnosis. Tan et al. investigated the performance of several representative one-class classifiers, considering common evaluation indices for machine learning, sample compositions of the training dataset, distribution of misclassified samples, and tolerance to contaminated data \citep{tan2020comparative}. This study is of great significance for providing decision support for applying one-class classifiers to other marine machinery systems. However, the scenario of limited label fault data for mechanical systems in the marine domain needs to be considered. Tan et al. trained multiple standard multi-label classification models using only single-fault data to determine unknown simultaneous faults \citep{tan2021multi}. Tan et al. proposed a mechanical attenuation state assessment method based on a one-class support vector machine (OCSVM), using the decision values returned by the trained model that only requires normal data and a small amount of labeled decay data to estimate the attenuation degree and main attenuation direction  \citep{tan2019one}. These methods have good performance under the closed set assumption of limited labeled data, while the situation where the fault class is unknown is common in fault diagnosis of marine mechanical systems.

\subsection{Problem definition}




Open-set fault diagnosis aims to identify both known and previously unseen fault types during testing. Let the labeled training dataset be defined as 

\begin{align}
D_l = \{(x_1^l, y_1^l), (x_2^l, y_2^l), \dots, (x_{n_l}^l, y_{n_l}^l)\},
\end{align}
where $x_i^l \in \mathbb{R}^m$ is an input feature vector representing $m$ measured variables, and $y_i^l \in \{1, 2, \dots, K\}$ is the corresponding class label from one of $K$ known fault classes.

The test set is denoted by
\begin{align}
D_u = \{(x_1^u, y_1^u), (x_2^u, y_2^u), \dots, (x_{n_u}^u, y_{n_u}^u)\},
\end{align}
with $n_u$ unlabeled samples drawn from both the known fault classes and an additional, unknown fault class indexed as $(K+1)$.


Traditional methods often train a classifier using only $D_l$, assuming that samples from unknown faults will fall outside the distribution of known classes. In practice, however, unknown and known faults may exhibit overlapping feature characteristics, making them difficult to separate using purely discriminative boundaries. This leads to misclassification and reduced diagnostic performance.

To address this, a subset of test samples $D_s \subset D_u$ can be selectively identified as potentially informative unknowns. Incorporating $D_s$ into the model training process, for example through pseudo-labeling or distribution refinement, can improve the model’s ability to distinguish unknown faults while maintaining performance on known classes \citep{sun2024conditional}.


\subsection{Graph representation}

The inherent structured knowledge of ship machinery systems includes topological connections, interaction strength, and signal propagation paths between sensors, which can help improve the accuracy, anti-interference ability, and interpretability of fault diagnosis. Generally, traditional fault diagnosis methods treat these multi-source sensor data as independent vectors, resulting in ineffective capture and utilization of structured knowledge.

The graph provides a powerful representation framework to address this issue, with the advantage of explicitly encoding the complex relationships between sensors. The operational status of the system is monitored by a set of sensors deployed at critical locations, forming the natural network from which structured graphical data can be obtained. The graph can be represented as $G=(V,E)$ with a node set $V$ and edge set $E$. Node $v_i \ in V $ is the sensor in systems, and edge $e_{i,j}=(v_i,v_j)\in E$ denotes the correlation between node ${v_i}$ and $v_j$. The adjacency matrix $A$ is a Boolean representation of the topological connections in the entire graph, and its element $a_ {i, j} $ takes a value of 1 when there is a direct connection relationship between node ${v_i}$ and $v_j$. The weight matrix $W$ assigns weights to the strength of connection relationships based on the adjacency matrix, and its elements $w_ {i, j} $ measure the strength of influence between nodes.

\subsection{Graph Convolutional Network}
In this study, we adopt the Graph Convolutional Network (GCN) based on Chebyshev polynomial approximation as the feature extractor, following \citet{li2022emerging}. Given a graph signal $x\in \mathbb{R}^m$, the spectral graph convolution operation is defined as:
\begin{align}
(x*_\mathcal{G}g)=U((U^Tx)\odot(U^Tg))=Ug_\theta U^Tx
\end{align}
where $*_\mathcal{G}$ is the graph convolution operator, $\odot$ is the element-wise Hadamard product, and $g_\theta = \mathrm{diag}(\theta)$ is a filter parameterized by $\theta$. The graph Fourier basis $U$ is the matrix of eigenvectors of the normalized graph Laplacian $L=I_m-D^{-1/2}AD^{-1/2}$. $A$ and $D$ are the adjacency matrix and degree matrix of the input graph respectively, and $I_m$ is an identity matrix.

To make the convolution operation more efficient and localized in space, Chebyshev polynomials is used to approximate the filter $g_theta$ and derived graph convolution of ChebyNet, which can be denoted as:
\begin{align}
g_\theta = \sum_{K-1}^{k=0}\theta_kT_k(\tilde{\Lambda})
\end{align}
\begin{align}
h=Ug_{\theta}U^Tx=\sum_{K-1}^{k=0}\theta_kT_k(\tilde{L})x
\end{align}
where $K$ is the order of the Chebyshev polynomials, and $K$ is set to 2 to simplify the convolution operation here. $\Lambda$ represents the diagonal matrix of eigenvalues of $L$, and $\tilde{\Lambda}=2\Lambda /\lambda_{max}-I_m$ is the rescale eigenvalue matrix. $\tilde{L}=2L/\lambda_{max}-I_n$ is the rescale Laplacian matrix and $T_k{\tilde{L}}=UT(\tilde{\Lambda})U^T$ denotes the Chebyshev polynomials.

\section{Open set-based fault diagnosis framework }
\label{sec:methodology}
\subsection{Pipeline}

\begin{figure}[h]
\centering
\includegraphics[width=\hsize]{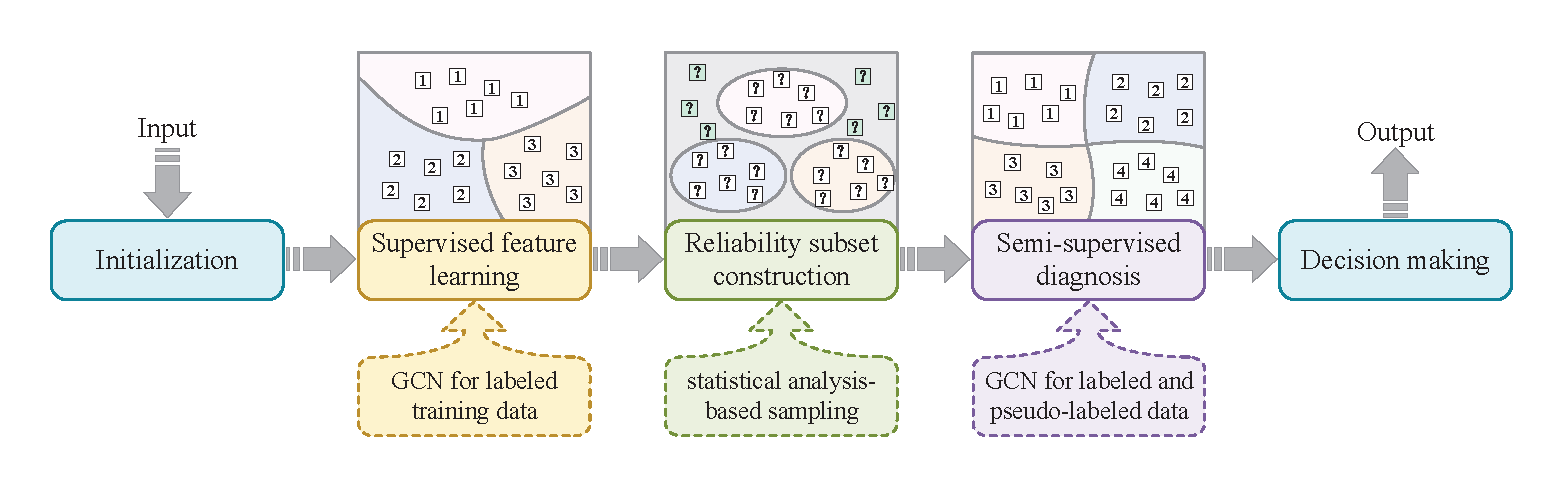}
\caption{Pipeline of the proposed fault diagnosis method.}  
\label{prfigure}
\end{figure}

The pipeline of the proposed open-set oriented fault diagnosis framework is shown in Figure \ref{prfigure}. The purpose of supervised feature learning is to extract discriminative features to distinguish known classes. The reliability subset construction process aims to collect test samples excluded from known classes to form an additional training set. Semi-supervised diagnosis implements accurate classification tasks for known and unknown samples.

\begin{figure}[h]
\centering
\includegraphics[width=\hsize]{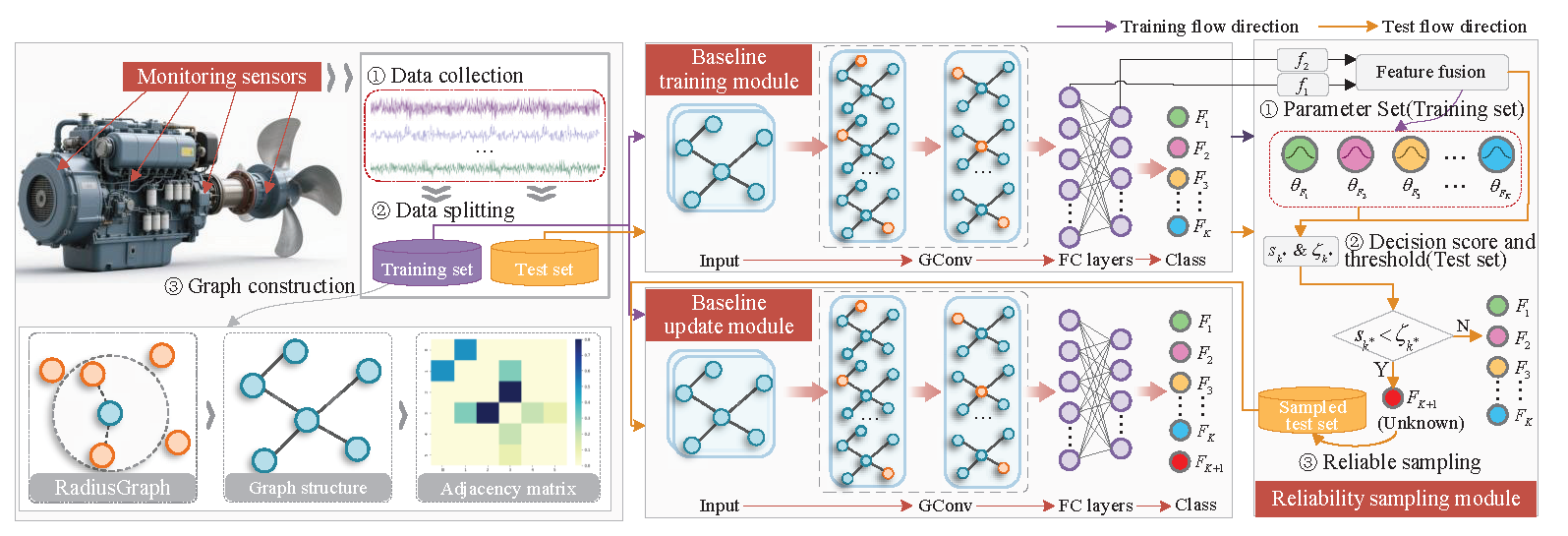}
\caption{The proposed fault diagnosis framework.}
\label{ffigure}
\end{figure}

Specifically, the framework of the method proposed in this article is shown in Figure \ref{ffigure}. In the initialization phase, graph data is constructed from time series collected by multiple sensors, and the edge weights between its nodes are used to reflect the relationship strength to consider the interdependence between data. The supervised feature learning model $M_0$ extracts features from the training graph data $D_l$ using graph convolutional layers and achieves accurate recognition of $K$ classes. The decision space for reliability sampling is formed by the fusion of output features from multiple fully connected layers of $M_0$. For each class $k \in \{1, 2, \dots, K\}$, a sampling parameter set $\theta_k$ is computed based on its training features in the decision space. The test samples from $D_u$ whose decision score $s_ {k ^ *}$ is less than the threshold $\zeta_ {k ^ *}$ are selected to form a relatively reliable subset $D_s$. Finally in the semi-supervised diagnosis process, the diagnosis model $M_1$ is trained according to the union set $D_l \cup D_s$ for making predictions on the test dataset $D_u$. The training of model $M_1$ ends when the number of iterations reaches the preset value, and the labels assigned to the test set $D_u$ by $M_1$ are used as the final prediction.

\subsection{Supervised feature learning}
Different sensors installed in ship propulsion systems for health status monitoring can be organized into structured graph data reflecting spatial information.  Given the graph $G=(V, E)$, the edge weight $w_{i,j}$ between node $v_i$ and its neighbor $v_j$ can be calculated by the Gaussian kernel weight function:

\begin{align}
w_{i,j}=\left\{
\begin{aligned}
	& exp(-\frac{d^2_{i,j}}{\sigma^2}),\, i\neq j \, \text{and} \, exp(-\frac{d^2_{i,j}}{\sigma^2}) \geq \varepsilon\\
	& 0, \, \text{otherwise}\\
\end{aligned}
\right.
\end{align}
where $d_{i,j}$ refers to the Euclidean distance between the $i$-th and $j$-th sensor signals in the training set. $\sigma^2$ and $\varepsilon$ represent thresholds that control the distribution and sparsity of the weight matrix $W=[w_{i,j}]_{m\times m}$, assigned to 10 and 0.5, respectively. Thus, each observation sample can be considered as a graphical representation of sensor measurements that are not independent but are linked by pairwise connection. 

The feature learning model $M_0$ uses GCN for fault diagnosis, with the labeled training set $D_l$ as input. The discriminant features of $K$ known fault classes are extracted via several graph convolution layers, and then fed into the Softmax classifier for prediction. For multi-class graph classification, the cross-entropy (CE) loss is utilized as the objective function:
\begin{align}
L_{CE}=-\frac{1}{N}\sum_{i=1}^N y_ilog(\hat{y}_i)
\end{align}
where $N$ takes the value of $n_l$ to improve the prediction accuracy of the model $M_0$ for $K$ known fault classes.

\subsection{Reliability subset construction}  

In this part, given the trained model $M_0$, a reliable pseudo-labeled subset $D_s$ is selected from the unlabeled test set $D_u$ based on multi-layer fusion features. Here, only the test samples whose discrimination scores are rejected by the known classes and have consistent predictions in the fusion feature space are considered as relatively reliable samples. Specifically, the reliability sample sampling approach is implemented in the following two steps: discrimination score calculation and consistent prediction sampling. 

For reliable subset collection, samples belonging to unknown classes need to be accurately excluded. Two perspectives are combined to construct the discriminant score function for unknown detection: (1) the distance between samples within known classes; (2) the distance between samples of unknown classes and samples of known classes. From the first perspective, the features extracted by the model $M_0$ make the samples of the same class closer in spatial distribution. From the second perspective, since the samples of unknown classes have not been exposed to the model $M_0$ in advance, they have low similarity with the features of known classes.  Based on this, the discriminant score function for unknown detection is constructed to achieve the recognition of samples from unknown classes. In the following, the definition of the discriminant score function is introduced in detail.

The model $M_0$ only learns discriminative features to distinguish known classes involved in training, resulting in other details being ignored in high-level features. Therefore, the multi-level features of $M_0$ are fused to form a new feature space to improve the recognition ability of unknown samples. The features output by multiple fully connected layers are fused to represent and separate unknown classes. The output feature vectors $z_i^l,z_i^{l+1},\dots,z_i^{l+c}$ from the $c$ fully connected layers of $M_0$ are integrated into the following joint representation:
\begin{align}
z_i=\text{CONCAT}\{z_i^l,z_i^{l+1},\dots,z_i^{l+c}\}
\end{align}
where $\text{CONCAT}\{\cdot\}$ operation concatenate $[z_i^l,z_i^{l+1},\dots,z_i^{l+c}]$ along feature dimension.

Considering a feature vector $z_i\in \mathbb{R}^d$, a discriminant analysis will assign $z_i$ to the fault class $F_k$ according to the following rule:

\begin{align}
z_i \in F_{k^*}, \text{if } k^*
&=
\mathop{\arg\max}_{k=1,\cdots,K}\frac{1}{\sum^K_{k=1} \textbf{e}\left(g_{k}(z_i) - g_{k}(z_i)\right)}
\end{align}

where $g_{k}(z_i)$ can be calculated as:
\begin{align}
g_{k}(z_i)=-\frac{1}{2}&\left(z_i-{\mu}_{k}\right)^{T} {\Sigma}_{k} ^{-1} \left(z_i-{\mu}_{k}\right) +\ln \left(P(F_k)\right)-\frac{1}{2} \ln \left( \left|  {\Sigma}_{k} \right| \right)\nonumber
\label{qda1}
\end{align}
where $\mu_k$and $\Sigma_k$ form the parameter set $\theta_{F_k}$ of the $F_k$ class, representing the mean vector and covariance matrix of the $F_k$ class samples, respectively. Generally, these parameters are estimated from data. $P(F_k)$ represents the prior probability of the class $F_k$, which can be obtained through expert knowledge.

To achieve unknown detection, statistical analysis is introduced, so that the $K+1$th sub-region is added to the feature space corresponding to the unknown observations that do not belong to or are far away from the training data. Based on the statistical test, the boundaries of known fault classes are constrained by statistical thresholds:
\begin{align}
g_{k^{*}}(z_i) \ge \frac{1}{2}L_{k^{*}}+ \tau_{k^{*}}
\end{align}
where $L_{k^*}$ is the control limit of the quadratic statistic given a significance level $\alpha$ and based on its statistical distribution. $L_{k^*}$ is given by \citep{ding2010application}:
\begin{align}
L_{k^*} = \frac{d \left( n_{k^*}^2-1\right) }{n_{k^*}\left( n_{k^*}-d\right) }F_\alpha (n_{k^*},n_{k^*}-d), \tau_{k^{*}}= \ln \left(P(F_{k^{*}})\left|{\Sigma}_{k^{*}} \right| ^{-\frac{1}{2}} \right) 
\end{align}
where $F_{\alpha}$ is the Fisher distribution and $\alpha$ refers to the confidence level. A lower $\alpha$ value leads to the shrinkage of the known class boundaries, classifying more known samples with small intra-class distances into the same class.

Since the distribution of unknown samples is far away from the known classes, the following statistical threshold is formed to achieve exclusion:
\begin{align}
\zeta_{k^*}= \frac{1}{\sum^K_{k=1} \textbf{e}(g_{k}(z_i) - \frac{1}{2}L_{k^{*}} -\tau_{k^{*}})}
\end{align}
The unlabeled sample $x_i^u$ in the test set $D_u$ first obtain the feature representation $z_i$ in the fused feature space of the model $M_0$. The discriminant score $s_{k^*}(z_i)$ of $z_i$ can be calculated according to the discriminant score function, and there is also a corresponding statistical threshold $\zeta_{k^*}$. If the discriminant score $s_{k^*}(z_i)$ is less than the statistical threshold $\zeta_{k^*}$, the sample $x_i^u$ will be assigned to the reliable set $D_s$ with pseudo-label $K+1$.

Regarding the consistent prediction sampling step,  for each sample in $D_p $, we search for its $C$ nearest neighbors from the set $D_u $ in the feature space based on Euclidean distances. Then, the test samples with pseudo labels consistent with $C/2$ neighbors in $D_p $ are retained, while the other samples are removed. These retained samples are identified as relatively reliable samples, and the subset formed by them is denoted as $D_s$.

\subsection{Semi-supervised diagnosis}
In this part, the training set $D_l $ with $K$ known classes and the test subset $D_s$ with pseudo-label $K+1$ are used together to train the semi-supervised diagnosis model $M_1$. Compared with the feature learning model $M_0$, the model $M_1$ adds the $K+1$th output, which has the ability to alleviate the overlap between known and unknown feature distributions. The complete implementation of SOFD-GCN for open set fault diagnosis of ship machinery systems using GCN as $M_0$ and $M_1$ models is shown in Algorithm 1.
\begin{table}[!h]
\centering
\label{TOSFD_algorithm}
\begin{tabular}{@{}p{\textwidth}@{}}
	\toprule
	\textbf{Algorithm 1:} SOFD-GCN for open-set fault diagnosis of marine machinery systems \\ 
	\midrule
	\begin{minipage}{\linewidth}
		\begin{algorithmic}[1]
			\State \textbf{Input:} labeled training set $D_l$ with $K$ known fault classes, unlabeled test set $D_u$ with known and unknown classes
			\State \textbf{Stage 1:} supervised feature learning
			\State\hspace*{1.2em} Data normalization.
			\State\hspace*{1.2em} Construct the graph with sensor measurements as nodes and weight matrix $W$.
			\State\hspace*{1.2em} Train GCN model $M_0$ with $D_l$.
			\State \textbf{Stage 2:} reliability subset construction
			\For{$x_i$ in $D_u$}
			\State Feature extraction: $z_i \leftarrow \left[ z_i^l,z_i^{l+1},\dots,z_i^{l+c} \right]_{\mathrm{Concat}}$.
			\State Calculate the discriminant score $s_{k^*}(z_i)$ of $x_i$.
			\State Calculate the statistical threshold $\zeta_{k^*}$ of $x_i$.
			\If{$s_{k^*}(z_i) < \zeta_{k^*}$}
			\State Put $x_i$ into the set $D_p$ with pseudo-label $K+1$.
			\EndIf
			\EndFor
			\For{$x_i$ in $D_p$}
			\State Count the number $n_p$ of $C$ nearest neighbors with label $K+1$ in the feature space.
			\If{$n_p > 2/C$}
			\State Put $x_i$ into the set $D_s$ with pseudo-label $K+1$.
			\EndIf
			\EndFor		
			\State \textbf{Stage 3:} semi-supervised diagnosis
			\State\hspace*{1.2em} Construct GCN model $M_1$ with $D_l$ and $D_s$.
			\State \textbf{Output:} diagnosis results of test set $D_u$ by updated GCN.
		\end{algorithmic}
	\end{minipage} \\
	\bottomrule
\end{tabular}
\end{table}

\section{Case study}
\label{sec:casestudy}

\subsection{Experimental Setup}
\subsubsection{Implementation details}
1800 data samples of each condition of the system were randomly selected, of which 70\% were used for training and 30\% for testing. To  demonstrate the effectiveness of the proposed method, sensor measurements collected at all nine different ship speeds are used for validation. The experiment is conducted with Windows 10 system, written in Python 3.9, and the experimental framework is Pytorch. For the model $M_0$, the structure of GCN is set to 32-32-32, and the structure of the fully connected layer in the classifier is 64-16-3 at speeds 1 and 2, and 64-8-3 at other speeds. The learning rate and batch size are set to 1e-5 and 64. We use Adam as the optimizer, and the training epoch is set to 100. During reliability sampling, the confidence level $\alpha$ is set to 0.01 and the number of nearest neighbors $C$ is set to 6.Subsequently, the labeled training set and pseudo-labeled test subset are used to train the model $M_1$, which has the same structure as $M_0$ except that an additional output is added to indicate the unknown class.
\subsubsection{Evaluation metrics}
We employ the following four metrics to evaluate the performance of the proposed SOFD-GCN and compared methods:

\textbullet\ U-recall: the unknown recall (U-recall) is a typical evaluation metric of the proportion of unknown samples in the test set that are correctly identified in open-set classification.

\textbullet\ ACC: the top-1 accuracy (ACC) measures the accuracy of correctly assigning all known samples in the test set to the corresponding known classes in closed-set classification..

\textbullet\ macro-F1: the macro-F1 score is a comprehensive performance evaluation indicator that integrates the evaluation of the classification accuracy on known classes and the detection ability of unknown classes of methods.

\textbullet\ confusion matrix: the confusion matrix is one of the most core evaluation indicators in the field of fault diagnosis, which visualizes the detailed classification results.

\subsection{Experimental results}
\subsubsection{Diagnosis results}

\textbf{Comparison with closed-set diagnosis methods.} In the work of \cite{tan2020comparative}, diagnosis results of four fault class at speed 5 were demonstrated using multiple one-class classifiers including one-class support vector machine (OCSVM), support vector data description (SVDD), global k-nearest neighbors (GKNN), local outlier factor (LOF), isolation forest (IForest) and angle-based outlier detection (ABOD). For test data of the four fault classes, ABOD achieved the highest ACC value of 0.9825, followed by OCSVM with a value of 0.965. The work in \cite{tan2019one} extends OCSVM to scenarios with limited labeled data, achieving prediction accuracy as high as 1 with only a few labeled samples at speeds 3 to 9. As shown in Table \ref{COM}, the proposed SOSFD-GCN achieves an ACC value of over 0.99 for samples of known fault classes at speeds 1 to 8, and a slightly lower ACC of 0.9778 at speed 9. It can be seen that the classification performance of the proposed SOFD-GCN for known classes is better than the above closed-set diagnosis methods.

\textbf{Comparison with open-set diagnosis methods.} In the experiments, six open set detection methods are compared with the proposed SOFD-GCN from the perspective of known and unknown fault classification for open-set diagnosis. OCSVM is extended to an open set diagnosis method as a continuation and improvement of the work of Tan et al \citep{tan2019one}. Specifically, an OCSVM model is established for each known fault class, and samples that do not belong to any class are excluded as unknown classes. Variants of discriminant analysis methods UQDA and UKFD \citep{lou2022novel} in the field of industrial system fault diagnosis are used as baseline comparison algorithms. Some classic open set detection algorithms such as Softmax, Openmax\citep{bendale2016towards} and OVRN-CD\citep{jang2022collective} are also used for comparison, and their baseline network structures are consistent with model $M_0$ of this paper. The classwise rejection threshold of Softmax is set to 0.9, and the threshold of OVRN-CD ensures that 95\% of the training samples are successfully identified as known.

\begin{table}[!h]
\centering
\caption{Evaluation results of different methods on open-set fault diagnosis task.}
\label{COM}
\scalebox{0.8}{
	\begin{tabular}{p{40pt}p{50pt}p{50pt}p{50pt}p{50pt}p{50pt}p{50pt}p{50pt}p{50pt}}
		\hline
		& Method & OCSVM & UQDA & UKFD & Softmax & Openmax & OVRN-CD & SOFD-GCN  \\	\hline
		\multirow{3}{*}{Speed 1} & U-recall & 0.9222 & 0.9111 & 0.8963 & 0.3926 & 0.4556 & 0.5556 & \textbf{0.9574} \\
		& ACC & 0.8006 & 0.8741 & 0.7827 & 0.9574 & 0.8253 & 0.9685 & \textbf{0.9963} \\
		& Macro-F1 & 0.8330 & 0.8878 & 0.8222 & 0.8043 & 0.7351 & 0.8602 & \textbf{0.9866} \\
		\hline
		\multirow{3}{*}{Speed 2} & U-recall & 0.7556 & 0.9389 & 0.9685 & 0.3185 & 0.8296 & 0.5130 & \textbf{1} \\
		& ACC & 0.8444 & 0.8031 & 0.8377 & 0.9870 & 0.9525 & 0.9623 & \textbf{0.9920} \\
		& Macro-F1 & 0.8274 & 0.8461 & 0.8771 & 0.7975 & 0.8419 & 0.8421 & \textbf{0.9940} \\
		\hline
		\multirow{3}{*}{Speed 3} & U-recall & 0.9648 & 0.7574 & 0.8889 & 0.0648 & 0.4500 & \textbf{0.9981} & 0.9019 \\
		& ACC & 0.8907 & 0.8494 & 0.9469 & \textbf{0.9994} & 0.9790 & 0.9519 & 0.9944 \\
		& Macro-F1 & 0.8970 & 0.8333 & 0.9322 & 0.6990 & 0.8364 & 0.9640 & \textbf{0.9712} \\
		\hline
		\multirow{3}{*}{Speed 4} & U-recall & \textbf{1} & \textbf{1} & \textbf{1} & 0.1500 & 0.4500 & \textbf{1} & 0.9926 \\
		& ACC & 0.7957 & 0.6037 & 0.8556 & \textbf{0.9981} & 0.9790 & 0.9222 & 0.9944 \\
		& Macro-F1 & 0.8512 & 0.6873 & 0.8937 & 0.7394 & 0.8364 & 0.9429 & \textbf{0.9940} \\
		\hline
		\multirow{3}{*}{Speed 5} & U-recall & \textbf{1} & 0.9259 & \textbf{1} & 0.4315 & 0.1722 & \textbf{1} & \textbf{1} \\
		& ACC & 0.9049 & 0.8475 & 0.8741 & \textbf{0.9981} & 0.9389 & 0.9191 & 0.9957 \\
		& Macro-F1 & 0.9299 & 0.8733 & 0.9085 & 0.8400 & 0.7046 & 0.9407 & \textbf{0.9968} \\
		\hline
		\multirow{3}{*}{Speed 6} & U-recall & 0.9796 & 0.8833 & \textbf{1} & 0.3426 & 0.2000 & \textbf{1} & \textbf{1} \\
		& ACC & 0.8710  & 0.8543 & 0.8463 & \textbf{0.9963} & 0.9265 & 0.9056 & 0.9944 \\
		& Macro-F1 & 0.8988 & 0.8676 & 0.8892 & 0.8117 & 0.7126 & 0.9309 & \textbf{0.9958} \\
		\hline
		\multirow{3}{*}{Speed 7} & U-recall & \textbf{1}  & 0.9296 & \textbf{1} & 0.1352 & 0.2556 & \textbf{1} & \textbf{1} \\
		& ACC & 0.8099 & 0.7457 & 0.7667 & \textbf{0.9957} & 0.9179 & 0.9074 & 0.9944 \\
		& Macro-F1 & 0.8574 & 0.8042 & 0.8298 & 0.7296 & 0.7312 & 0.9322 & \textbf{0.9958} \\
		\hline
		\multirow{3}{*}{Speed 8} & U-recall & 0.9222 & 0.8222 & 0.9852 & 0.3222 & 0.1963 & \textbf{1} & \textbf{1} \\
		& ACC & 0.9019 & 0.7549 & 0.8710 & 0.9920 & 0.9099 & 0.9216 & \textbf{0.9951} \\
		& Macro-F1 & 0.9089 & 0.7833 & 0.9019 & 0.8020 & 0.7018 & 0.9424 & \textbf{0.9963} \\
		\hline
		\multirow{3}{*}{Speed 9} & U-recall & 0.8149 & 0.8981 & 0.9167 & 0.1778 & 0.1741 & 0.5796 & \textbf{1} \\
		& ACC & 0.9006 & 0.8167 & 0.9056 & \textbf{0.9920} & 0.9457 & 0.9475 & 0.9778 \\
		& Macro-F1 & 0.8785& 0.8455 & 0.9111 & 0.7402 & 0.7091 & 0.8530 & \textbf{0.9832} \\
		\hline
	\end{tabular}
}
\end{table}

As seen from Table \ref{COM}, compared with other open-set diagnosis methods, the performances of the proposed SOFD-GCN are significantly improved on the propulsion system dataset with unknown fault samples. The five compared methods apply thresholds calculated from the labeled training set to the output discriminant scores, where test samples that do not exceed the threshold are identified as unknown classes. OCSVM, UQDA and UKFD perform better in unknown sample detection than known sample classification, which is reflected in higher U-recall values and lower ACC values. Deep learning methods including Softmax, Openmax, and OVRN-CD tend to overfit to the known class samples involved in training, resulting in superior performance in closed-set recognition. Softmax and Openmax are prone to outputting overconfident classification probabilities of some unknown class samples and misclassifying them into known classes, thus failing to effectively reject unknown samples. The introduction of collective decision-making OVRN-CD alleviates the inherent over-generalization of deep learning classifiers, and unknown samples are given high confidence scores, which significantly improves the U-recall value at speeds 3, 4, 5, 6, 7, and 8. SOFD-GCN achieves the highest U-recall value at almost all speeds, indicating that the participation of the reliable unlabeled test subset effectively improves the discriminability of unknown samples. For closed-set fault diagnosis, SOFD-GCN has the highest accuracy at speeds 1 and 2, while the ACC values at other speeds are all above 0.95 and slightly lower than Softmax. The highest macro-F1 value of SOFD-GCN at all speeds illustrates its best overall performance in open-set fault diagnosis, which is capable of completing the two core tasks of known identification and unknown detection.

\begin{figure}[!h]
\centering
\includegraphics[width=\hsize]{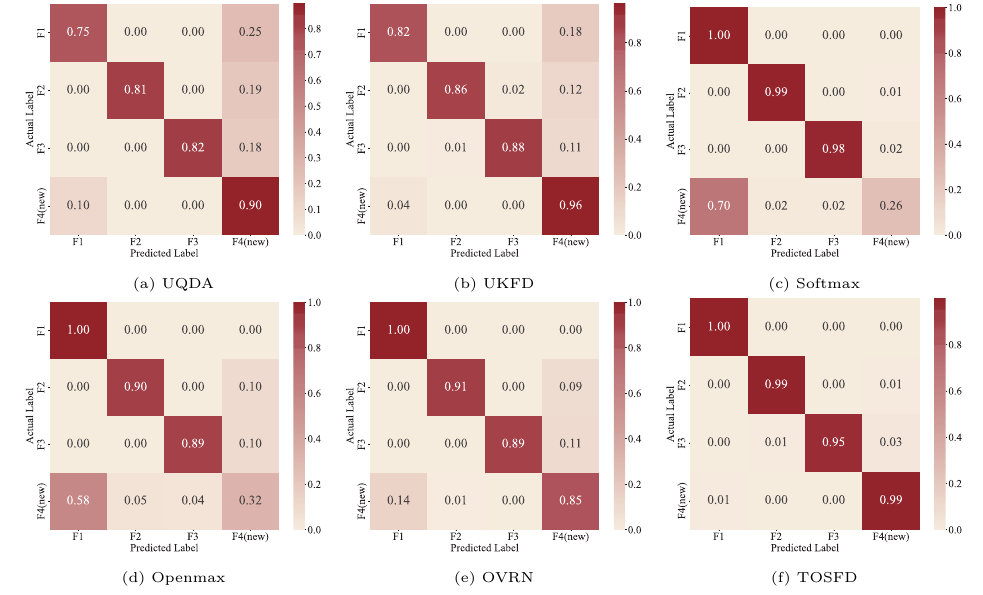}
\caption{Confusion matrices of the compared methods and the proposed SOFD-GCN at 9 speeds (New denotes the unknown fault).}
\label{confusion}
\end{figure}

To provide detailed information on the open diagnosis results of the proposed SOFD-GCN and compared methods, confusion matrices showing the distribution of all test samples at 9 speeds are given in Figure \ref{confusion}. UQDA and UKFD can accurately identify unknown samples, while also incorrectly excluding some samples from known classes due to the tightened statistical thresholds calculated based on the labeled training set. Softmax and Openmax are restricted by the closed set assumption, which makes the discriminant features for distinguishing unknown samples learned under unconstrained conditions, resulting in a large number of unknown samples being mistakenly attributed to Fault 1. OVRN-CD establishes a strict and compact decision boundary to reject more new samples as unknown classes, increasing the unknown detection rate to 0.85. Figure \ref{confusion}(f) shows that SOFD-GCN achieves impressive accuracies on known Fault 1, Fault 2 and the unknown fault with nearly 1, while the accuracy on Fault 3 is slightly lower at 0.95.Figure \ref{confusion}(f) shows that SOFD-GCN achieves impressive accuracies on known Fault 1, Fault 2 and the unknown fault with nearly 1, while the accuracy on Fault 3 is slightly lower at 0.95. The results in Figure \ref{confusion} illustrate that the proposed SOFD-GCN is able to simultaneously achieve accurate classification of known fault classes and effective detection of unknown fault samples.

\subsubsection{Feature Visualization}

\begin{figure}[!h]
\centering
\includegraphics[width=\hsize]{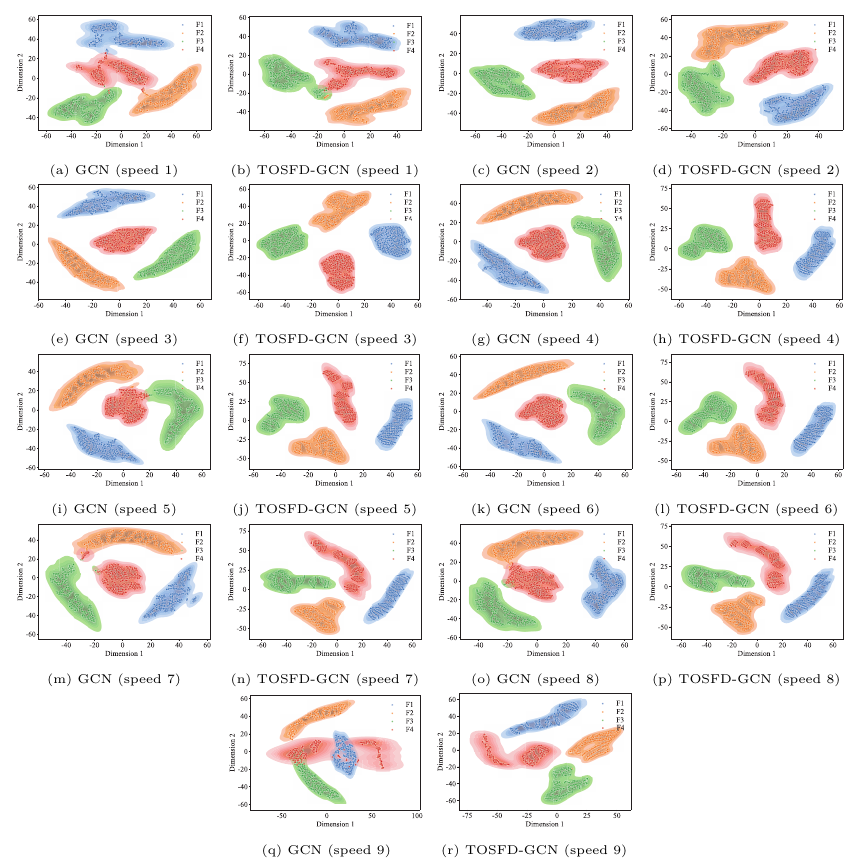}
\caption{T-SNE visualization of features extracted by model $M_0$ and after semi-supervised learning by $M_1$ for three known classes ($F_1$, $F_2$, $F_3$) and one unknown class ($F_4$) at 9 speeds.}.
\label{tsne}
\end{figure}

T-SNE is adopted to visualize the feature representations extracted from last fully connected layers of the feature learning model $M_0$ and the semi-supervised diagnosis model $M_1$ for all test samples at 9 different speeds. The visualization results are shown in Figure \ref{tsne}, where the four class samples are marked with different colors, with $F_4$ as the unknown class.

As can be seen in Figure \ref{tsne}, almost all samples of the same class form compact representations in the feature space. The features extracted by $M_0$ have similar performance to $M_1$  at speeds 2, 3, 4, and 6, achieving inter-class separation of the four classes. At speeds 1, 5, 6, 8, and 9, the unknown class features (red) are overlapped with the known class features, indicating that the features learned by the model $M_0$  from known samples are not sufficient to exclude unknown samples. Obviously, SOFD-GCN significantly reduces the overlapping area, making the unknown class features independently distributed away from known classes. Benefiting from the introduction of the reliable pseudo-labeled test subset, the extracted features simultaneously learn discriminative features of known and unknown classes, thus solving the class under-representation problem.

\subsection{Further discussion}
\subsubsection{Ablation study}

\textbf{Ablation study on the GCN-based supervised feature learning for fault diagnosis.} The proposed method employs GCN as the  feature learning model for feature learning and fault classification. GCN can effectively utilize the correlation between nodes for information transmission and reasoning, and is suitable for application in the field of fault diagnosis with graph structure characteristics. The applicability of GCN in UAV sensor fault diagnosis is verified by comparing it with SOFD-CNN, which uses CNN instead of GCN as models $M_0$ and $M_1$. In this experiment, the convolution kernel size of CNN is set to $1\times 2$, and other network structure parameters are consistent with GCN. The results are shown in Table \ref{A1}. GCN as the feature learning and classification model performs better than CNN under the proposed SOFD framework. Specifically, compared to SOFD-CNN, SOFD-GCN has slightly lower recognition accuracy for unknown fault samples, but higher classification accuracy for known fault samples. Overall, SOFD-GCN performs better in UAV fault diagnosis, as reflected in the Marco-F1 value of 0.9904. SOFD-CNN tends to correctly assign more unknown samples, as CNN is sensitive to distribution shift due to missing topological information and overfitting of local features. Therefore, the GCN architecture that integrates physical topology information can be a reliable approach.

\textbf{Ablation study on the proposed feature fusion strategy for unknown detection.} The implementation of the proposed reliability sampling for pseudo-labeled test samples is based on the fused feature space of the model $M_0$. We evaluate the impact of feature space representation in supervised feature learning process on fault diagnosis. Specifically, SOFD-GCN is modified into the following two methods for comparison:

(i) SOFD-GCN without supervised feature learning: here, the model $M_0$ is removed and then the reliability sampling of the test samples is performed directly in the original data space instead of the extracted deep feature space.

(ii) SOFD-GCN without feature fusion: here, the feature fusion strategy is not executed, and only the last layer output of $M_0$ is used as the feature space to sample the test samples.

\begin{table}[!h]
\centering
\caption{Comparison of three metrics for ablation study on feature fusion strategy.}
\label{A1}
\begin{tabular}{p{200pt}p{40pt}p{40pt}p{50pt}}
	\hline
	Method &   UR &  ACC  & Macro-F1 \\ \hline
	SOFD-CNN   &   \textbf{0.9924}  & 0.9831 & 0.9855 \\  \hline
	SOFD-GCN without supervised training  &   0.9621  & 0.9841 & 0.9786 \\ 
	SOFD-GCN without feature fusion    &   0.9311  & 0.9926 & 0.9770 \\   \hline
	SOFD-GCN   &   0.9835  & \textbf{0.9927} & \textbf{0.9904} \\
	\hline
\end{tabular}
\end{table}

The diagnosis results are provided in Table \ref{A1}, which are the averages of the indicators UR, ACC and Macro-F1 at nine speeds. Since the model $M_0$ works under the closed set assumption, the features learned by SOFD-GCN without feature fusion focus on known fault classes, resulting in the loss of discriminative features that can exclude unknown samples. SOFD-GCN without supervised training utilizes the complete information from the original data and thus obtains a higher UR value of 0.9621. However, due to the limitations of class separability in the original data space, it is inevitable that some test samples are assigned incorrect pseudo labels. Addressing this problem, fused feature space based reliability sampling strategy is proposed to improve the accuracy of fault diagnosis under open-set assumption. As can be seen from the table, SOFD-GCN performs best on all metrics, which shows that multiple hidden layers are capable of obtaining more complex fault representations. The introduction of the feature fusion strategy improves the separation between known and unknown classes, thereby facilitating the selection of reliable test samples with unknown pseudo labels.

\begin{figure}[!h]
\centering
\includegraphics[width=0.6\hsize]{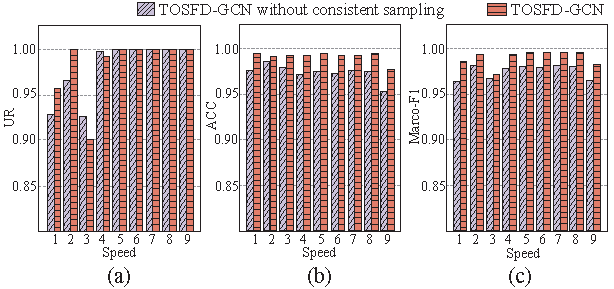}
\caption{Comparison of three metrics for ablation study on sampling approach.}
\label{A123}
\end{figure}

\textbf{Ablation study on the proposed reliability subset construction approach.} The proposed unknown detection method based on statistical analysis builds boundaries for known classes and selects test samples that are consistent with their neighbors' pseudo-labels to recombine into the reliable subset. Consistency sampling is ignored, and test samples assigned with unknown pseudo-labels are directly used in the semi-supervised diagnosis model. For three evaluation metrics at nine speeds, the comparison between SOFD-GCN and SOFD-GCN without consistent sampling is given in Figure \ref{A123}. Note that at most speeds, SOFD-GCN performs better than the version without consistent sampling. Consistent sampling further improves the selection of reliable test samples, which alleviates the problem of incorrect selection of known class samples for unknown detection only through discriminant scores.

\subsubsection{Hyperparameter sensitivity}
\begin{figure}[!h]  
\centering
\captionsetup[subfigure]{font=tiny}
\begin{subfigure}[b]{0.4\textwidth}
	\centering
	\includegraphics[width=\textwidth]{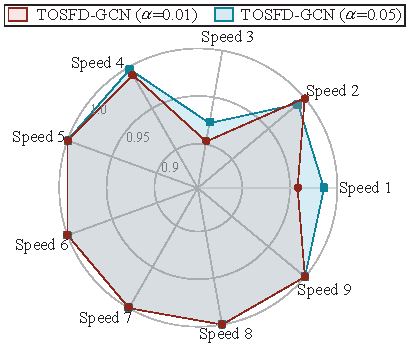}
	\caption{UR for comparison}
\end{subfigure}
\begin{subfigure}[b]{0.4\textwidth}
	\centering
	\includegraphics[width=\textwidth]{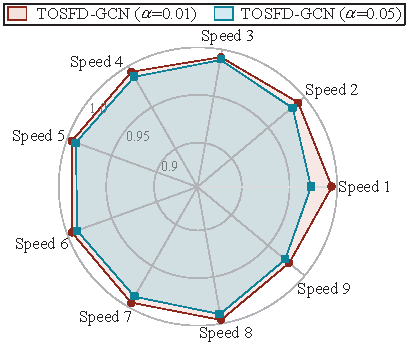}
	\caption{ACC for comparison}
\end{subfigure}
\caption{Comparison of diagnostic accuracy for different $\alpha$ values.}
\label{alpha}
\end{figure}

The value of the  significance level $\alpha$ affects the selection of the test subset involved in the training in the reliability sampling module. An experiment was added to evaluate the effect of fault diagnosis on the propulsion system at nine speeds with $\alpha$ set to 0.05. Two metrics include UR and ACC of SOFD-GCN are shown in Figure \ref{alpha}, where $\alpha$ is set to 0.01 and 0.05 for comparison. It can be observed that SOFD-GCN with $\alpha=0.05$ has slightly lower ACC values for known classes at all speeds. Moreover, SOFD-GCN with $\alpha=0.05$ has higher UR values at peeds 1, 3, and 4, demonstrating higher unknown detection capabilities. The reason is that a higher value of $\alpha$ would lead to shrinked boundaries of known classes, and then more exclusions. When selecting relatively reliable samples, a higher $\alpha$ causes more test samples that originally belong to known classes to be incorrectly assigned pseudo-labels of unknown classes. These mislabeled test samples with similar discriminative features to the training samples may confuse the semi-supervised model $M_1$, resulting in a decrease in the classification performance of known samples. 

\section{Conclusion}
\label{sec:conclusion}
In this paper, a realistic fault diagnosis scheme for marine mechanical systems called a semi-supervised open-set fault diagnosis (SOFD) is proposed. The GCN-based supervised feature learning model is first built and learned with labeled training samples, and then multi-layer fused feature representations are obtained for unlabeled test data. A reliable pseudo-labeled test subset is selected combining the exclusion criteria based on statistical analysis and consistent sampling, and finally used together with the training set to complete the training of the semi-supervised diagnosis model. Experiments are conducted on a dataset generated from a real data validated simulator of a vessel to verify the effectiveness of the proposed method. Specifically, compared with five classic open-set fault diagnosis methods, SOFD-GCN achieves the best overall performance with macro-F1 scores exceeding 0.97 at all nine speeds, demonstrating its superiority in real shipping industry applications. The graph structure is fundamental to the SOFD framework. The integration of GCN allows for effective modeling of the relationships among sensors, facilitating robust feature learning with limited annotations. Furthermore, the graph-based representations enhance both feature discrimination and the reliability of the pseudo-label selection, which is instrumental for the success of the semi-supervised open-set diagnosis.

Although the proposed SOFD framework shows its effectiveness in fault diagnosis under the open set assumption, there are still some challenges that need to be addressed. In the actual application of marine mechanical systems, the distribution of training data and test data is inconsistent due to the complex operating characteristics and working environment, which is not considered in this paper. Therefore, further improving the performance of SOFD to adapt to the data distribution difference will become our future research work.

\bibliographystyle{model}
\bibliography{ref}

\begin{thebibliography}{37}
\expandafter\ifx\csname natexlab\endcsname\relax\def\natexlab#1{#1}\fi
\providecommand{\bibinfo}[2]{#2}
\ifx\xfnm\relax \def\xfnm[#1]{\unskip,\space#1}\fi
\bibitem[{Xu et~al.(2023)Xu, Lin, and Ye}]{xu2023fault}
\bibinfo{author}{X.~Xu}, \bibinfo{author}{Y.~Lin}, \bibinfo{author}{C.~Ye},
\newblock \bibinfo{title}{Fault diagnosis of marine machinery via an
  intelligent data-driven framework},
\newblock \bibinfo{journal}{Ocean Engineering} \bibinfo{volume}{289}
  (\bibinfo{year}{2023}) \bibinfo{pages}{116302}.
\bibitem[{Lazakis et~al.(2018)Lazakis, Raptodimos, and
  Varelas}]{lazakis2018predicting}
\bibinfo{author}{I.~Lazakis}, \bibinfo{author}{Y.~Raptodimos},
  \bibinfo{author}{T.~Varelas},
\newblock \bibinfo{title}{Predicting ship machinery system condition through
  analytical reliability tools and artificial neural networks},
\newblock \bibinfo{journal}{Ocean Engineering} \bibinfo{volume}{152}
  (\bibinfo{year}{2018}) \bibinfo{pages}{404--415}.
\bibitem[{Raptodimos and Lazakis(2020)}]{raptodimos2020application}
\bibinfo{author}{Y.~Raptodimos}, \bibinfo{author}{I.~Lazakis},
\newblock \bibinfo{title}{Application of narx neural network for predicting
  marine engine performance parameters},
\newblock \bibinfo{journal}{Ships and Offshore Structures} \bibinfo{volume}{15}
  (\bibinfo{year}{2020}) \bibinfo{pages}{443--452}.
\bibitem[{Zhao et~al.(2023)Zhao, Zhang, Mao, and Jiang}]{zhao2023variational}
\bibinfo{author}{N.~Zhao}, \bibinfo{author}{J.~Zhang},
  \bibinfo{author}{Z.~Mao}, \bibinfo{author}{Z.~Jiang},
\newblock \bibinfo{title}{Variational time--frequency adaptive decomposition of
  machine multi-impact vibration signals},
\newblock \bibinfo{journal}{Mechanical Systems and Signal Processing}
  \bibinfo{volume}{189} (\bibinfo{year}{2023}) \bibinfo{pages}{110084}.
\bibitem[{Li et~al.(2023)Li, Jiao, and Geng}]{li2023refined}
\bibinfo{author}{Y.~Li}, \bibinfo{author}{S.~Jiao}, \bibinfo{author}{B.~Geng},
\newblock \bibinfo{title}{Refined composite multiscale fluctuation-based
  dispersion lempel--ziv complexity for signal analysis},
\newblock \bibinfo{journal}{ISA transactions} \bibinfo{volume}{133}
  (\bibinfo{year}{2023}) \bibinfo{pages}{273--284}.
\bibitem[{Tan et~al.(2021)Tan, Zhang, Tian, Jiang, Guo, Wang, and
  Lin}]{tan2021multi}
\bibinfo{author}{Y.~Tan}, \bibinfo{author}{J.~Zhang},
  \bibinfo{author}{H.~Tian}, \bibinfo{author}{D.~Jiang},
  \bibinfo{author}{L.~Guo}, \bibinfo{author}{G.~Wang},
  \bibinfo{author}{Y.~Lin},
\newblock \bibinfo{title}{Multi-label classification for simultaneous fault
  diagnosis of marine machinery: a comparative study},
\newblock \bibinfo{journal}{Ocean Engineering} \bibinfo{volume}{239}
  (\bibinfo{year}{2021}) \bibinfo{pages}{109723}.
\bibitem[{Vanem and Brands{\ae}ter(2021)}]{vanem2021unsupervised}
\bibinfo{author}{E.~Vanem}, \bibinfo{author}{A.~Brands{\ae}ter},
\newblock \bibinfo{title}{Unsupervised anomaly detection based on clustering
  methods and sensor data on a marine diesel engine},
\newblock \bibinfo{journal}{Journal of Marine Engineering \& Technology}
  \bibinfo{volume}{20} (\bibinfo{year}{2021}) \bibinfo{pages}{217--234}.
\bibitem[{Lv et~al.(2024)Lv, Yang, Li, Liu, and Li}]{lv2024fault}
\bibinfo{author}{Y.~Lv}, \bibinfo{author}{X.~Yang}, \bibinfo{author}{Y.~Li},
  \bibinfo{author}{J.~Liu}, \bibinfo{author}{S.~Li},
\newblock \bibinfo{title}{Fault detection and diagnosis of marine diesel
  engines: A systematic review},
\newblock \bibinfo{journal}{Ocean Engineering} \bibinfo{volume}{294}
  (\bibinfo{year}{2024}) \bibinfo{pages}{116798}.
\bibitem[{Li et~al.(2025)Li, Atoui, and Li}]{li2025dual}
\bibinfo{author}{G.~Li}, \bibinfo{author}{M.~A. Atoui},
  \bibinfo{author}{X.~Li},
\newblock \bibinfo{title}{Dual adversarial and contrastive network for
  single-source domain generalization in fault diagnosis},
\newblock \bibinfo{journal}{Advanced Engineering Informatics}
  \bibinfo{volume}{65} (\bibinfo{year}{2025}) \bibinfo{pages}{103140}.
\bibitem[{Han et~al.(2020)Han, Li, Skulstad, Skjong, and Zhang}]{han2020deep}
\bibinfo{author}{P.~Han}, \bibinfo{author}{G.~Li},
  \bibinfo{author}{R.~Skulstad}, \bibinfo{author}{S.~Skjong},
  \bibinfo{author}{H.~Zhang},
\newblock \bibinfo{title}{A deep learning approach to detect and isolate
  thruster failures for dynamically positioned vessels using motion data},
\newblock \bibinfo{journal}{IEEE Transactions on Instrumentation and
  Measurement} \bibinfo{volume}{70} (\bibinfo{year}{2020})
  \bibinfo{pages}{1--11}.
\bibitem[{Shahid et~al.(2022)Shahid, Ko, and Kwon}]{shahid2022real}
\bibinfo{author}{S.~M. Shahid}, \bibinfo{author}{S.~Ko},
  \bibinfo{author}{S.~Kwon},
\newblock \bibinfo{title}{Real-time abnormality detection and classification in
  diesel engine operations with convolutional neural network},
\newblock \bibinfo{journal}{Expert Systems with Applications}
  \bibinfo{volume}{192} (\bibinfo{year}{2022}) \bibinfo{pages}{116233}.
\bibitem[{Wang et~al.(2021)Wang, Chen, and Guan}]{wang2021random}
\bibinfo{author}{R.~Wang}, \bibinfo{author}{H.~Chen},
  \bibinfo{author}{C.~Guan},
\newblock \bibinfo{title}{Random convolutional neural network structure: An
  intelligent health monitoring scheme for diesel engines},
\newblock \bibinfo{journal}{Measurement} \bibinfo{volume}{171}
  (\bibinfo{year}{2021}) \bibinfo{pages}{108786}.
\bibitem[{Gao et~al.(2024)Gao, Xu, Zhang, Liu, and Chang}]{gao2024marine}
\bibinfo{author}{B.~Gao}, \bibinfo{author}{J.~Xu}, \bibinfo{author}{Z.~Zhang},
  \bibinfo{author}{Y.~Liu}, \bibinfo{author}{X.~Chang},
\newblock \bibinfo{title}{Marine diesel engine piston ring fault diagnosis
  based on lstm and improved beluga whale optimization},
\newblock \bibinfo{journal}{Alexandria Engineering Journal}
  \bibinfo{volume}{109} (\bibinfo{year}{2024}) \bibinfo{pages}{213--228}.
\bibitem[{Han et~al.(2021)Han, Ellefsen, Li, {\AE}s{\o}y, and
  Zhang}]{han2021faulta}
\bibinfo{author}{P.~Han}, \bibinfo{author}{A.~L. Ellefsen},
  \bibinfo{author}{G.~Li}, \bibinfo{author}{V.~{\AE}s{\o}y},
  \bibinfo{author}{H.~Zhang},
\newblock \bibinfo{title}{Fault prognostics using lstm networks: application to
  marine diesel engine},
\newblock \bibinfo{journal}{IEEE Sensors Journal} \bibinfo{volume}{21}
  (\bibinfo{year}{2021}) \bibinfo{pages}{25986--25994}.
\bibitem[{Ellefsen et~al.(2020)Ellefsen, Han, Cheng, Holmeset, {\AE}s{\o}y, and
  Zhang}]{ellefsen2020online}
\bibinfo{author}{A.~L. Ellefsen}, \bibinfo{author}{P.~Han},
  \bibinfo{author}{X.~Cheng}, \bibinfo{author}{F.~T. Holmeset},
  \bibinfo{author}{V.~{\AE}s{\o}y}, \bibinfo{author}{H.~Zhang},
\newblock \bibinfo{title}{Online fault detection in autonomous ferries: Using
  fault-type independent spectral anomaly detection},
\newblock \bibinfo{journal}{IEEE Transactions on instrumentation and
  measurement} \bibinfo{volume}{69} (\bibinfo{year}{2020})
  \bibinfo{pages}{8216--8225}.
\bibitem[{Velasco-Gallego and Lazakis(2022)}]{velasco2022radis}
\bibinfo{author}{C.~Velasco-Gallego}, \bibinfo{author}{I.~Lazakis},
\newblock \bibinfo{title}{Radis: A real-time anomaly detection intelligent
  system for fault diagnosis of marine machinery},
\newblock \bibinfo{journal}{Expert Systems with Applications}
  \bibinfo{volume}{204} (\bibinfo{year}{2022}) \bibinfo{pages}{117634}.
\bibitem[{Han et~al.(2021)Han, Ellefsen, Li, Holmeset, and
  Zhang}]{han2021faultb}
\bibinfo{author}{P.~Han}, \bibinfo{author}{A.~L. Ellefsen},
  \bibinfo{author}{G.~Li}, \bibinfo{author}{F.~T. Holmeset},
  \bibinfo{author}{H.~Zhang},
\newblock \bibinfo{title}{Fault detection with lstm-based variational
  autoencoder for maritime components},
\newblock \bibinfo{journal}{IEEE Sensors Journal} \bibinfo{volume}{21}
  (\bibinfo{year}{2021}) \bibinfo{pages}{21903--21912}.
\bibitem[{Wang et~al.(2023)Wang, Chen, and Guan}]{wang2023self}
\bibinfo{author}{R.~Wang}, \bibinfo{author}{H.~Chen},
  \bibinfo{author}{C.~Guan},
\newblock \bibinfo{title}{A self-supervised contrastive learning framework with
  the nearest neighbors matching for the fault diagnosis of marine machinery},
\newblock \bibinfo{journal}{Ocean Engineering} \bibinfo{volume}{270}
  (\bibinfo{year}{2023}) \bibinfo{pages}{113437}.
\bibitem[{Wang et~al.(2022)Wang, Chen, and Guan}]{wang2022dpgcn}
\bibinfo{author}{R.~Wang}, \bibinfo{author}{H.~Chen},
  \bibinfo{author}{C.~Guan},
\newblock \bibinfo{title}{Dpgcn model: A novel fault diagnosis method for
  marine diesel engines based on imbalanced datasets},
\newblock \bibinfo{journal}{IEEE Transactions on Instrumentation and
  Measurement} \bibinfo{volume}{72} (\bibinfo{year}{2022})
  \bibinfo{pages}{1--11}.
\bibitem[{Zhou et~al.(2024)Zhou, Fang, Li, Wang, and
  Kung}]{zhou2024contrastive}
\bibinfo{author}{Y.~Zhou}, \bibinfo{author}{S.~Fang}, \bibinfo{author}{S.~Li},
  \bibinfo{author}{B.~Wang}, \bibinfo{author}{S.-Y. Kung},
\newblock \bibinfo{title}{Contrastive learning based open-set recognition with
  unknown score},
\newblock \bibinfo{journal}{Knowledge-Based Systems} \bibinfo{volume}{296}
  (\bibinfo{year}{2024}) \bibinfo{pages}{111926}.
\bibitem[{Guo et~al.(2024)Guo, Zhang, Sun, and Wang}]{guo2024universal}
\bibinfo{author}{Y.~Guo}, \bibinfo{author}{J.~Zhang}, \bibinfo{author}{B.~Sun},
  \bibinfo{author}{Y.~Wang},
\newblock \bibinfo{title}{A universal fault diagnosis framework for marine
  machinery based on domain adaptation},
\newblock \bibinfo{journal}{Ocean Engineering} \bibinfo{volume}{302}
  (\bibinfo{year}{2024}) \bibinfo{pages}{117729}.
\bibitem[{Wang et~al.(2024)Wang, Gao, Li, and Gao}]{wang2024self}
\bibinfo{author}{L.~Wang}, \bibinfo{author}{Y.~Gao}, \bibinfo{author}{X.~Li},
  \bibinfo{author}{L.~Gao},
\newblock \bibinfo{title}{Self-supervised-enabled open-set cross-domain fault
  diagnosis method for rotating machinery},
\newblock \bibinfo{journal}{IEEE Transactions on Industrial Informatics}
  (\bibinfo{year}{2024}).
\bibitem[{Zhao and Shen(2022)}]{zhao2022adaptive}
\bibinfo{author}{C.~Zhao}, \bibinfo{author}{W.~Shen},
\newblock \bibinfo{title}{Adaptive open set domain generalization network:
  Learning to diagnose unknown faults under unknown working conditions},
\newblock \bibinfo{journal}{Reliability Engineering \& System Safety}
  \bibinfo{volume}{226} (\bibinfo{year}{2022}) \bibinfo{pages}{108672}.
\bibitem[{Sun et~al.(2023)Sun, Yang, and Lin}]{9953149}
\bibinfo{author}{H.~Sun}, \bibinfo{author}{B.~Yang}, \bibinfo{author}{S.~Lin},
\newblock \bibinfo{title}{An open set diagnosis method for rolling bearing
  faults based on prototype and reconstructed integrated network},
\newblock \bibinfo{journal}{IEEE Transactions on Instrumentation and
  Measurement} \bibinfo{volume}{72} (\bibinfo{year}{2023})
  \bibinfo{pages}{1--10}.
\bibitem[{Liu et~al.(2023)Liu, Zhang, Pan, Zhang, Hong, Wang, Wang, and
  Miao}]{10198472}
\bibinfo{author}{Z.~Liu}, \bibinfo{author}{X.~Zhang}, \bibinfo{author}{J.~Pan},
  \bibinfo{author}{X.~Zhang}, \bibinfo{author}{W.~Hong},
  \bibinfo{author}{Z.~Wang}, \bibinfo{author}{Z.~Wang},
  \bibinfo{author}{Y.~Miao},
\newblock \bibinfo{title}{Similar or unknown fault mode detection of aircraft
  fuel pump using transfer learning with subdomain adaption},
\newblock \bibinfo{journal}{IEEE Transactions on Instrumentation and
  Measurement} \bibinfo{volume}{72} (\bibinfo{year}{2023})
  \bibinfo{pages}{1--11}.
\bibitem[{Fu et~al.(2023)Fu, Bi, Han, Zhang, Liu, Zhao, and Hu}]{10214203}
\bibinfo{author}{R.~Fu}, \bibinfo{author}{Y.~Bi}, \bibinfo{author}{G.~Han},
  \bibinfo{author}{X.~Zhang}, \bibinfo{author}{L.~Liu},
  \bibinfo{author}{L.~Zhao}, \bibinfo{author}{B.~Hu},
\newblock \bibinfo{title}{Magva: An open-set fault diagnosis model based on
  multi-hop attentive graph variational autoencoder for autonomous vehicles},
\newblock \bibinfo{journal}{IEEE Transactions on Intelligent Transportation
  Systems} \bibinfo{volume}{24} (\bibinfo{year}{2023})
  \bibinfo{pages}{14873--14889}.
\bibitem[{Yang et~al.(2021)Yang, Wei, Sun, Li, Zhou, Xiong, and
  Yang}]{yang2021s2osc}
\bibinfo{author}{Y.~Yang}, \bibinfo{author}{H.~Wei}, \bibinfo{author}{Z.-Q.
  Sun}, \bibinfo{author}{G.-Y. Li}, \bibinfo{author}{Y.~Zhou},
  \bibinfo{author}{H.~Xiong}, \bibinfo{author}{J.~Yang},
\newblock \bibinfo{title}{S2osc: A holistic semi-supervised approach for open
  set classification},
\newblock \bibinfo{journal}{ACM Transactions on Knowledge Discovery from Data
  (TKDD)} \bibinfo{volume}{16} (\bibinfo{year}{2021}) \bibinfo{pages}{1--27}.
\bibitem[{Sun and Dong(2024)}]{sun2024conditional}
\bibinfo{author}{J.~Sun}, \bibinfo{author}{Q.~Dong},
\newblock \bibinfo{title}{Conditional feature generation for transductive
  open-set recognition via dual-space consistent sampling},
\newblock \bibinfo{journal}{Pattern Recognition} \bibinfo{volume}{146}
  (\bibinfo{year}{2024}) \bibinfo{pages}{110046}.
\bibitem[{Cipollini et~al.(2018)Cipollini, Oneto, Coraddu, Murphy, and
  Anguita}]{cipollini2018condition}
\bibinfo{author}{F.~Cipollini}, \bibinfo{author}{L.~Oneto},
  \bibinfo{author}{A.~Coraddu}, \bibinfo{author}{A.~J. Murphy},
  \bibinfo{author}{D.~Anguita},
\newblock \bibinfo{title}{Condition-based maintenance of naval propulsion
  systems: Data analysis with minimal feedback},
\newblock \bibinfo{journal}{Reliability Engineering \& System Safety}
  \bibinfo{volume}{177} (\bibinfo{year}{2018}) \bibinfo{pages}{12--23}.
\bibitem[{Carrega et~al.(2019)Carrega, Cipollini, and
  Oneto}]{carrega2019simple}
\bibinfo{author}{A.~Carrega}, \bibinfo{author}{F.~Cipollini},
  \bibinfo{author}{L.~Oneto},
\newblock \bibinfo{title}{Simple continuous optimal regions of the space of
  data},
\newblock \bibinfo{journal}{Neurocomputing} \bibinfo{volume}{349}
  (\bibinfo{year}{2019}) \bibinfo{pages}{91--104}.
\bibitem[{Tan et~al.(2020)Tan, Tian, Jiang, Lin, and
  Zhang}]{tan2020comparative}
\bibinfo{author}{Y.~Tan}, \bibinfo{author}{H.~Tian},
  \bibinfo{author}{R.~Jiang}, \bibinfo{author}{Y.~Lin},
  \bibinfo{author}{J.~Zhang},
\newblock \bibinfo{title}{A comparative investigation of data-driven approaches
  based on one-class classifiers for condition monitoring of marine machinery
  system},
\newblock \bibinfo{journal}{Ocean Engineering} \bibinfo{volume}{201}
  (\bibinfo{year}{2020}) \bibinfo{pages}{107174}.
\bibitem[{Tan et~al.(2019)Tan, Niu, Tian, Hou, and Zhang}]{tan2019one}
\bibinfo{author}{Y.~Tan}, \bibinfo{author}{C.~Niu}, \bibinfo{author}{H.~Tian},
  \bibinfo{author}{L.~Hou}, \bibinfo{author}{J.~Zhang},
\newblock \bibinfo{title}{A one-class svm based approach for condition-based
  maintenance of a naval propulsion plant with limited labeled data},
\newblock \bibinfo{journal}{Ocean Engineering} \bibinfo{volume}{193}
  (\bibinfo{year}{2019}) \bibinfo{pages}{106592}.
\bibitem[{Li et~al.(2022)Li, Zhou, Li, Sun, Yan, and Chen}]{li2022emerging}
\bibinfo{author}{T.~Li}, \bibinfo{author}{Z.~Zhou}, \bibinfo{author}{S.~Li},
  \bibinfo{author}{C.~Sun}, \bibinfo{author}{R.~Yan},
  \bibinfo{author}{X.~Chen},
\newblock \bibinfo{title}{The emerging graph neural networks for intelligent
  fault diagnostics and prognostics: A guideline and a benchmark study},
\newblock \bibinfo{journal}{Mechanical Systems and Signal Processing}
  \bibinfo{volume}{168} (\bibinfo{year}{2022}) \bibinfo{pages}{108653}.
\bibitem[{Ding et~al.(2010)Ding, Zhang, Ding, Naik, Deng, and
  Gui}]{ding2010application}
\bibinfo{author}{S.~Ding}, \bibinfo{author}{P.~Zhang},
  \bibinfo{author}{E.~Ding}, \bibinfo{author}{A.~Naik},
  \bibinfo{author}{P.~Deng}, \bibinfo{author}{W.~Gui},
\newblock \bibinfo{title}{On the application of pca technique to fault
  diagnosis},
\newblock \bibinfo{journal}{Tsinghua Science and Technology}
  \bibinfo{volume}{15} (\bibinfo{year}{2010}) \bibinfo{pages}{138--144}.
\bibitem[{Lou et~al.(2022)Lou, Atoui, and Li}]{lou2022novel}
\bibinfo{author}{C.~Lou}, \bibinfo{author}{M.~A. Atoui},
  \bibinfo{author}{X.~Li},
\newblock \bibinfo{title}{Novel online discriminant analysis based schemes to
  deal with observations from known and new classes: Application to industrial
  systems},
\newblock \bibinfo{journal}{Engineering Applications of Artificial
  Intelligence} \bibinfo{volume}{111} (\bibinfo{year}{2022})
  \bibinfo{pages}{104811}.
\bibitem[{Bendale and Boult(2016)}]{bendale2016towards}
\bibinfo{author}{A.~Bendale}, \bibinfo{author}{T.~E. Boult},
\newblock \bibinfo{title}{Towards open set deep networks},
\newblock in: \bibinfo{booktitle}{Proceedings of the IEEE conference on
  computer vision and pattern recognition}, pp. \bibinfo{pages}{1563--1572}.
\bibitem[{Jang and Kim(2022)}]{jang2022collective}
\bibinfo{author}{J.~Jang}, \bibinfo{author}{C.~O. Kim},
\newblock \bibinfo{title}{Collective decision of one-vs-rest networks for
  open-set recognition},
\newblock \bibinfo{journal}{IEEE Transactions on Neural Networks and Learning
  Systems} \bibinfo{volume}{35} (\bibinfo{year}{2022})
  \bibinfo{pages}{2327--2338}.

\end{thebibliography}

\end{document}